\documentclass[preprint,12pt]{elsarticle}

\usepackage{amssymb}
\usepackage{bm}

\journal{arXiv}

\begin{document}

\begin{frontmatter}



\title{Theory of Hallucinations based on Equivariance}

\author[inst1]{Hisaichi Shibata}
\cortext[cor1]{Corresponding Author}
\ead{sh@g.ecc.u-tokyo.ac.jp}
\affiliation[inst1]{organization={Department of Radiology, The~University of Tokyo Hospital},
            addressline={7-3-1~Hongo}, 
            city={Bunkyo, Tokyo 113-8655},
            country={Japan}}
\begin{abstract}
This study aims to acquire knowledge for creating very large language models that are immune to hallucinations. Hallucinations in contemporary large language models are often attributed to a misunderstanding of real-world social relationships. Therefore, I hypothesize that very large language models capable of thoroughly grasping all these relationships will be free from hallucinations. Additionally, I propose that certain types of equivariant language models are adept at learning and understanding these relationships. Building on this, I have developed a specialized cross-entropy error function to create a hallucination scale for language models, which measures their extent of equivariance acquisition. Utilizing this scale, I tested language models for their ability to acquire character-level equivariance. In particular, I introduce and employ a novel technique based on T5 (Text To Text Transfer Transformer) that efficiently understands permuted input texts without the need for explicit dictionaries to convert token IDs (integers) to texts (strings). This T5 model demonstrated a moderate ability to acquire character-level equivariance. Additionally, I discovered scale laws that can aid in developing hallucination-free language models at the character level. This methodology can be extended to assess equivariance acquisition at the word level, paving the way for very large language models that can comprehensively understand relationships and, consequently, avoid hallucinations.
\end{abstract}




\end{frontmatter}


\section{Introduction}
"Gargling with stones and using the stream as a pillow."
This old proverb refers to an attitude of not admitting mistakes.
Analyzing the structure of this proverb reveals a misunderstanding in the relationship between action and object.
Namely, stones are not for gargling, and a stream is not for resting one's head.
Meanwhile, in the modern era, a similar phenomenon occurs even with the latest language models, such as ChatGPT.
In reality, when I quoted a dialogue spoken by a character in an anime and asked ChatGPT to identify who said it, it confidently mentioned the wrong character. Although this is a case of mistaking fictional characters, it is conceivable that similar mistakes can happen with real people, and this misidentification ultimately stems from misrecognizing relationships in the real world. This phenomenon, known as hallucination, poses the greatest challenge even for cutting-edge large-scale language models to overcome.

\begin{figure}[htbp]
    \centering
    \includegraphics[width=0.9\textwidth]{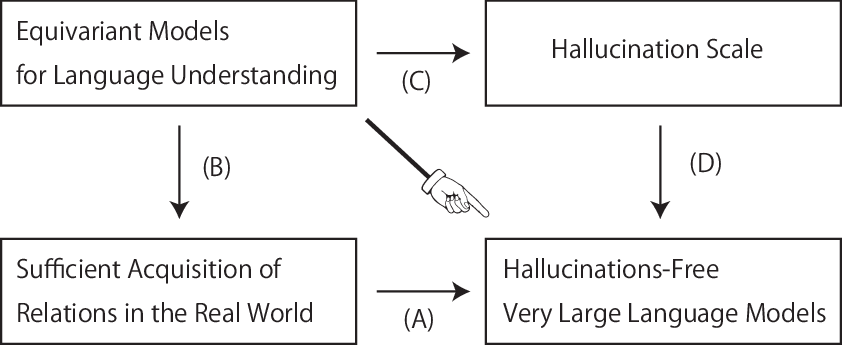}
    \caption{The goal of this study is to gain insights that contribute to the realization of a hallucination-free, very large language model (corresponding to the diagonal arrows).}
    \label{fig:goal}
\end{figure}

This study proposes a mathematical framework capable of handling such misinterpretations and develops a theory that can contribute to reducing hallucinations in cutting-edge large-scale language models (see Figure \ref{fig:goal}). Discussions on hallucinations not caused by misinterpretations of relationships are avoided. In this research, firstly, it hypothesizes that a model sufficiently acquiring relationships in the real world (as envisioned in a broad and detailed manner, almost logical but also probabilistic, as depicted in Figure 2a) can realize a hallucination-free, large-scale language model. Additionally, it argues that a model, in a sense equivariant, corresponds to its sufficient acquisition of relationships.
Furthermore, it proposes a Hallucination Scale, an evaluative measure quantifying the acquisition of such equivariance by the model. Lastly, although it is a toy model, it quantifies the intensity of actual hallucinations based on this Hallucination Scale. Particularly, the Hallucination Scale proposed in this research, grounded in the mathematical concept of equivariance, can investigate whether the language model correctly learns and infers all relationships among people, objects, concepts, and subjective experiences in the real world. It quantifies the strength of hallucinations in this sense (in this research, it is considered that the strength of hallucinations increases with the increase in misinterpretations of relationships in the real world). Thus, this paper contributes to the emergence of hallucination-free, large-scale language models.

\section{Theory of Hallucinations}

The driving scientific question of this study is, "What are the conditions for the disappearance of hallucinations?" It should be noted that while the theory proposed in this research could potentially be applicable to hallucinations occurring in humans, the focus henceforth will be solely on hallucinations arising in large-scale language models, not humans. The following sections will provide detailed explanations of arrows (A), (B), (C), and (D) as mentioned in Figure \ref{fig:goal}.

\subsection{Disappearance of hallucinations through the understanding of relationships (A)}

\begin{figure}[htbp]
    \centering
    \includegraphics[width=0.5\textwidth]{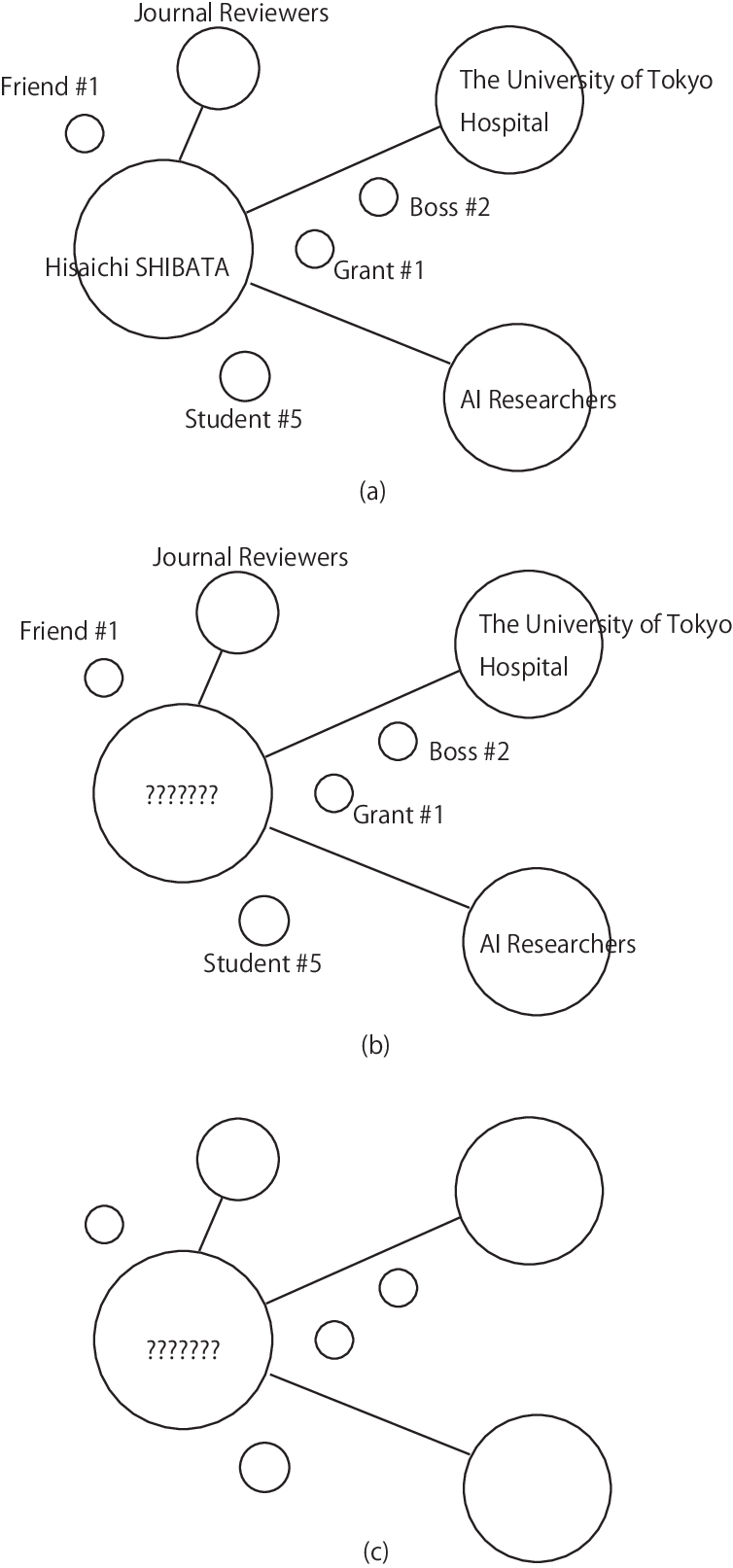}
    \caption{Examples of relationships in real society. The words inside the circle include not only names of people and objects but also actions and abstract concepts. Here, relationships are described as undirected graphs, but in reality, they also encompass logical relationships, probabilistic elements, and set relationships. (a) A part of the relationships in real society as seen by the author, (b) the method found in traditional language models (such as BERT \cite{devlin2018bert}) that estimates specific words from relationships (word gap-filling problem), and (c) the new method proposed in this study, which estimates all words (including the existence of people and objects and concepts) solely from relationships.}
    \label{fig:relationships}
\end{figure}

When processing natural language with digital computers and language models, it is common to convert strings (text) into sequences of token IDs, which are integer values, using a tokenizer. If the tokenizer's dictionary (the correspondence rules between strings and integer values) is damaged (as corresponding to Figure \ref{fig:relationships}c), it is known that recovering the dictionary is not easy (corresponding to recovering Figure \ref{fig:relationships}a from Figure \ref{fig:relationships}c). For example, if a natural sentence like "Apples are red and sweet." is passed through a tokenizer, it might output a sequence of token IDs like "3301, 2, 40, 3, 1, 3235, 10, 5, 7", and then suppose the tokenizer loses its dictionary. If a decoder mistakenly interprets "3301" as "Lemon", then "40, 3", which originally corresponded to "red", is statistically more likely to be interpreted as "yellow" (though difficult to prove conclusively, statistically red lemons are less plausible). Similarly, "3235", which corresponded to "sweet", may be statistically more likely to be interpreted as "sour" (again, a hard proof but sweet lemons are less plausible). This could lead to a chain reaction altering the original meaning of the text. However, there's also a possibility that no alternative method exists to recover a dictionary different from the original one, implying that the tokenizer's original dictionary can be uniquely recovered.

If the text provided during training or inference is extremely scarce, there could be immense possible decoding methods, thus increasing the likelihood of generating misinterpreted text (i.e., the emergence of hallucinations). However, as the volume of text increases, the relationships between token IDs (not just grammar but also the relative relationships of objects in real society as captured by the text) begin to restrict the possible methods of decoding. To summarize abstractly, if one sufficiently acquires the relationships in real society (here, the relative relationships among people, objects, abstract concepts, and any subjective experiences), then hallucinations (misinterpretations of relationships) are likely to disappear, a claim that is tautological but not necessarily obvious. Additionally, in practical terms, it is believed that having sufficiently long training time and a sufficiently large number of model parameters are prerequisites for the disappearance of hallucinations.

\subsection{Understanding of relationships and acquisition of equivariance (B)}
\label{subsec:equivariance_and_relationships}

\begin{figure}[htbp]
    \centering
    \includegraphics[width=\textwidth]{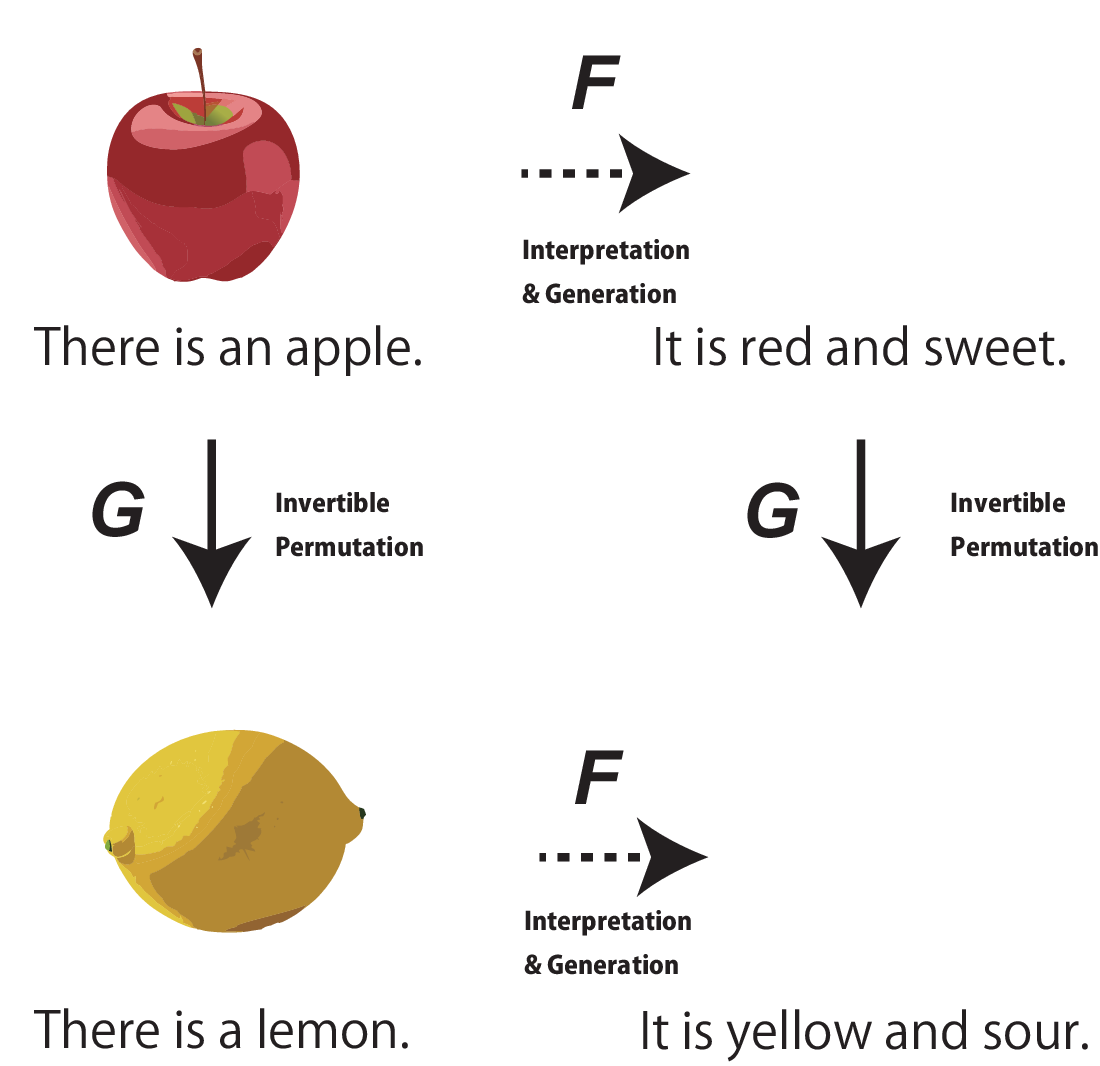}
    \caption{An example of equivariant models handled in this study (explained using objects)}
    \label{fig:equivariant_model_ex1}
\end{figure}

\begin{figure}[htbp]
    \centering
    \includegraphics[width=\textwidth]{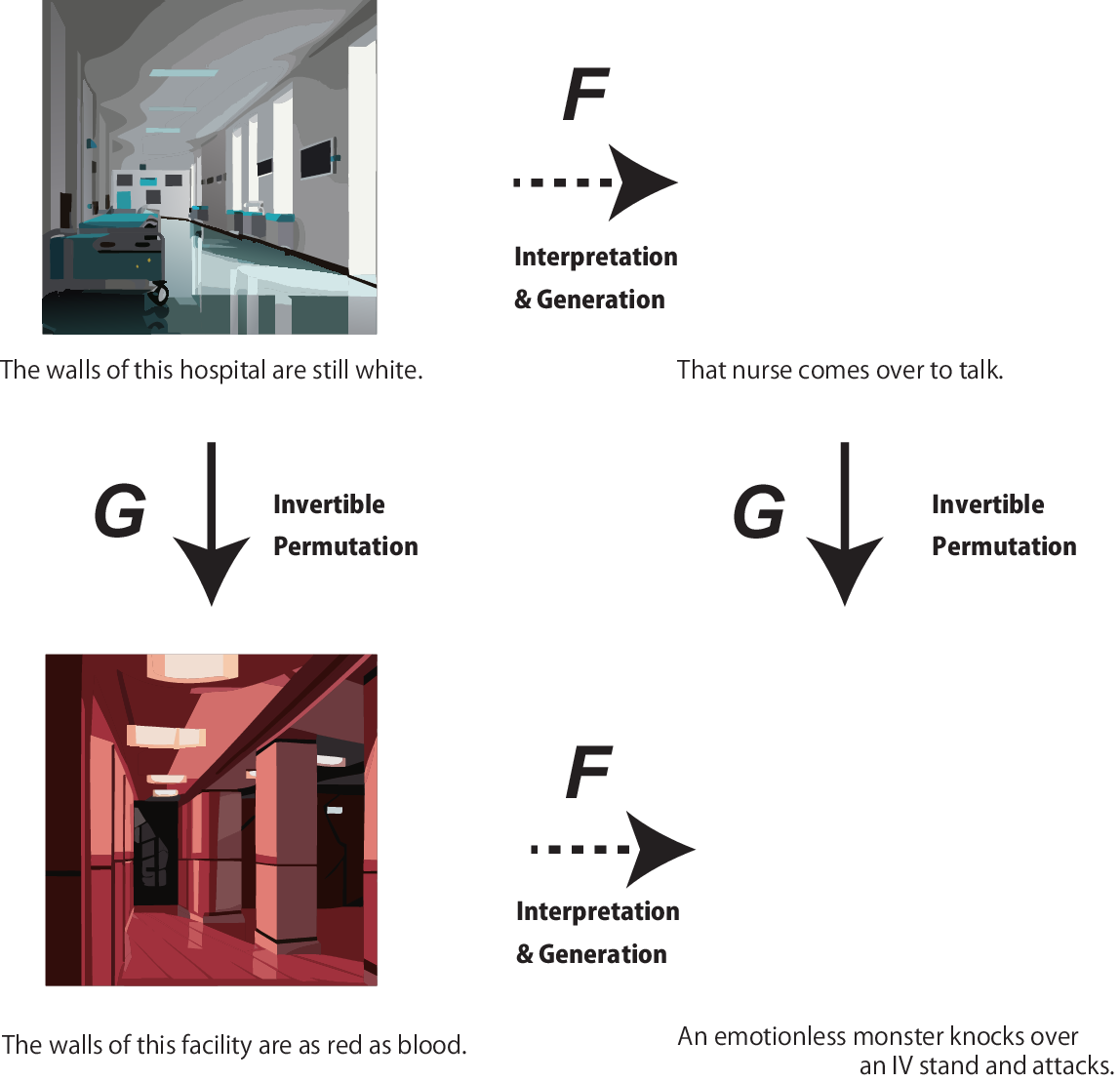}
    \caption{An example of equivariant models handled in this study (explained using fictional subjective experiences)}
    \label{fig:equivariant_model_ex2}
\end{figure}

In the following, I explain how the function of understanding the relationships between people and objects in real society can be abstractly represented using the mathematical tool of equivariance. Equivariance is an important concept that frequently appears in machine learning, and has recently been featured in Style-GAN-3 \cite{karras2021alias}. However, to aid the reader's understanding, I first explain equivariance in the context of image processing. For a model F' with equivariance (for example, think of a segmentation neural network), the output obtained by inputting an image to which operation G' (rotation) has been applied is the same as the result of applying operation G' to the output of the original image input to model F' (a model with rotational equivariance). On the other hand, the language model F with equivariance proposed in this study is defined as a model F whose interpretation results remain invariant between the interpretation of a natural sentence after its token ID representation has been consistently permuted (G) based on a certain invertible dictionary, and the results after an invertible permutation (G) is applied to the interpretation (application of language model F) of the pre-permutation natural sentence (see Figures \ref{fig:equivariant_model_ex1} and \ref{fig:equivariant_model_ex2}).

Expressed in mathematical terms, this situation can be represented as:

\begin{equation}
    \bm{F} \circ \bm{G} = \bm{G} \circ \bm{F},
\end{equation}
where this equation must hold for all input texts.

In models with this equivariance, the token IDs assigned to words representing all people, objects, etc., can be anything as long as there is no duplication (equivalent to a wild card). A language model that can consistently handle words like wild cards, in other words, can be said to understand the relationships of everything in the real world (real society), and this is a model that has acquired equivariance.

It is important to note that equivariance is distinctly different from invariance. If we fix the natural sentence and merely change the content (integer values) of the token ID assignment, the meaning of the natural sentence represented by the sequence of token IDs changes (note the change in the meaning of the text before and after the permutation G in Figures \ref{fig:equivariant_model_ex1} and \ref{fig:equivariant_model_ex2}). Therefore, the proposed language model is not a model with invariance to permutation operations. That is, for permutations G other than the identity permutation,
\begin{equation}
    \bm{F} \circ \bm{G} \not= \bm{F}.
\end{equation}

\subsection{Scales for assessing the ability to acquire equivariance (C)}
\label{subsec:scales}
When only a large number of token ID sequences representing text (such as "3301, 2, 40, 3, 1, 3235, 10, 5, 7") are provided, it is practically impossible for us humans to manually restore the corresponding dictionary (i.e., to restore words from token IDs).

On the other hand, if such a task (specifically, having the model output dictionary estimation results and calculating the cross-entropy error with a ground truth dictionary) could be automated through an algorithm, it would be a scale capable of fully automatically assessing the ability to acquire equivariance. Moreover, as discussed in Sub-Section \ref{subsec:equivariance_and_relationships}, there is a close relationship between the state of equivariance acquisition and the strength of hallucinations; therefore, this scale can assess the strength of hallucinations. Hence, in this study, I model the rules for restoring words (or phrases) from token ID sequences using neural networks. This problem is equivalent to deciphering a word substitution cipher (deliberately erasing the dictionary to verify if the language model has sufficiently acquired equivariance).

Conversely, language models like BERT \cite{devlin2018bert} learn the structure of text by, for example, hiding parts of a token ID sequence like "3301, 2, 40, 3, 1, 3235, 10, 5, 7" and making predictions. This is equivalent to predicting words with prior knowledge of all words except the ones to be estimated (refer to Figure \ref{fig:relationships}b), and compared to the method proposed in this paper, the difficulty of learning and inference is considered significantly lower. However, to rigorously assess and guarantee the absence of any misinterpretation of relationships in a language model, the method proposed in this study, where words are estimated from a state where all words including the ones to be estimated are unknown (Figure \ref{fig:relationships}c), is considered a more appropriate approach.

\subsection{Emergence of a hallucination-free, very large language model based on the hallucination scale (D)}
If a scale capable of quantifying the strength of hallucinations can be constructed, it increases the likelihood of developing a language model that is free from hallucinations, using this scale as an evaluation metric. Realistically, it is conceivable to discover scaling laws related to the strength of hallucinations and to quantify how many parameters, the number of tokenized texts for training, and how much training time are required for a model to sufficiently suppress hallucinations.
In particular, it has been considered a challenging topic to verify how much the frequency of hallucinations decreases when scaling up the amount of training text and increasing the number of constraints compared to traditional approaches. However, by adopting this proposed scale, it is believed that such verification can be mechanically performed.

\section{Related Works}
As previously explained, to confirm that a model can acquire equivariance, one can measure the robustness of the language model F against invertible permutations G of words (or characters or phrases). Invertible permutations have long been known in the field of cryptography as character substitution ciphers. Notably, the dancing men cipher in Arthur Conan Doyle's detective stories is a classic representation of a character substitution cipher. Here's the essence of character substitution ciphers: In such ciphers, for example, an invertible correspondence rule is established between the alphabet and space characters as follows:

Before permutation: abcdefghijklmnopqrstuvwxyz 

After permutation: mwt qovhdxizaulgcyresjkfpnb

An efficient analysis of character substitution ciphers often involves frequency analysis, comparing the statistical frequency of characters in the cipher text with the frequency of characters in natural sentences in real society. However, methods for deciphering character and word substitution ciphers that do not rely on frequency analysis have received less attention. The mechanical decipherment of character substitution ciphers by computers, which have traditionally been heuristically decoded by humans, is an academically intriguing topic. In this study, I achieve a qualitatively new method of decipherment using deep learning technology, specifically T5 \cite{raffel2020exploring,tay2021scale}, which can handle text transformation rules.

Aldarrad et al. \cite{aldarrab2021can} adopted a Seq2Seq model to attempt the automatic decryption of character substitution ciphers using machine learning, but they argued that frequency analysis was necessary to achieve sufficient analytical accuracy. Kambhatla et al. \cite{kambhatla2023decipherment} used the decoder block of the Transformer \cite{vaswani2017attention} to successfully achieve sufficient accuracy in the automatic decryption of character substitution ciphers through machine learning, but they did not discuss the extension to word substitution ciphers or their relationship to hallucinations.

\section{Numerical Experiments}
In Sub-Section \ref{subsec:scales}, I stated that the ability of a neural network to decipher word substitution ciphers could be used to determine the extent to which the network has acquired equivariance. However, the training cost for such a neural network is expected to be high. Therefore, within the scope of this paper, I discuss modeling character substitution ciphers, which are thought to have dramatically lower computational costs compared to word substitution ciphers, as a toy model.

In this study, I adopted the T5 models from HuggingFace's Transformers library: google/t5-efficient-tiny (16M parameters), google/t5-efficient-mini (30M parameters), and google/t5-efficient-small (60M parameters). For the test dataset, I used the first 4,096 lines from CC-100 (English). As training data, I extracted an additional 4,096*N lines from the same location (here, N is an integer value proportional to the size of the dataset, set at 1, 8, 16 in this study). All English texts were converted to lower camel case.
Punctuation was removed, and a token ID was assigned to spaces between words to learn word boundaries.
The number of input tokens was capped at 512. During training, (i) I created a new dictionary containing only 26 alphabetic characters, a space character, and BOS and EOS symbols. (ii) The English texts were encoded using this dictionary to obtain a sequence of token IDs. (iii) Next, the token IDs were substituted using a random but invertible substitution rule (specifically, using the permutation function of the numpy library). The substitution rule was changed for each data, including the test dataset. The model was trained with T5 to output the substitution rule (i.e., the dictionary itself), different from previous studies, using the substituted token ID sequence as input. The loss function was cross-entropy error for the token ID sequence, and T5 learned general rules for predicting output text from input text. The batch size was set at 16. I used the Adam optimizer with a learning rate of 0.0001. Additionally, I called a function to manage the vocabulary size of T5 and modified the vocabulary settings. All calculations in this paper were performed on a laptop equipped with a single NVIDIA-RTX-4090 Mobile, and training was continued for up to 100 epochs for all cases.

\section{Results}
\begin{figure}[htbp]
    \centering
    \includegraphics[width=0.9\textwidth]{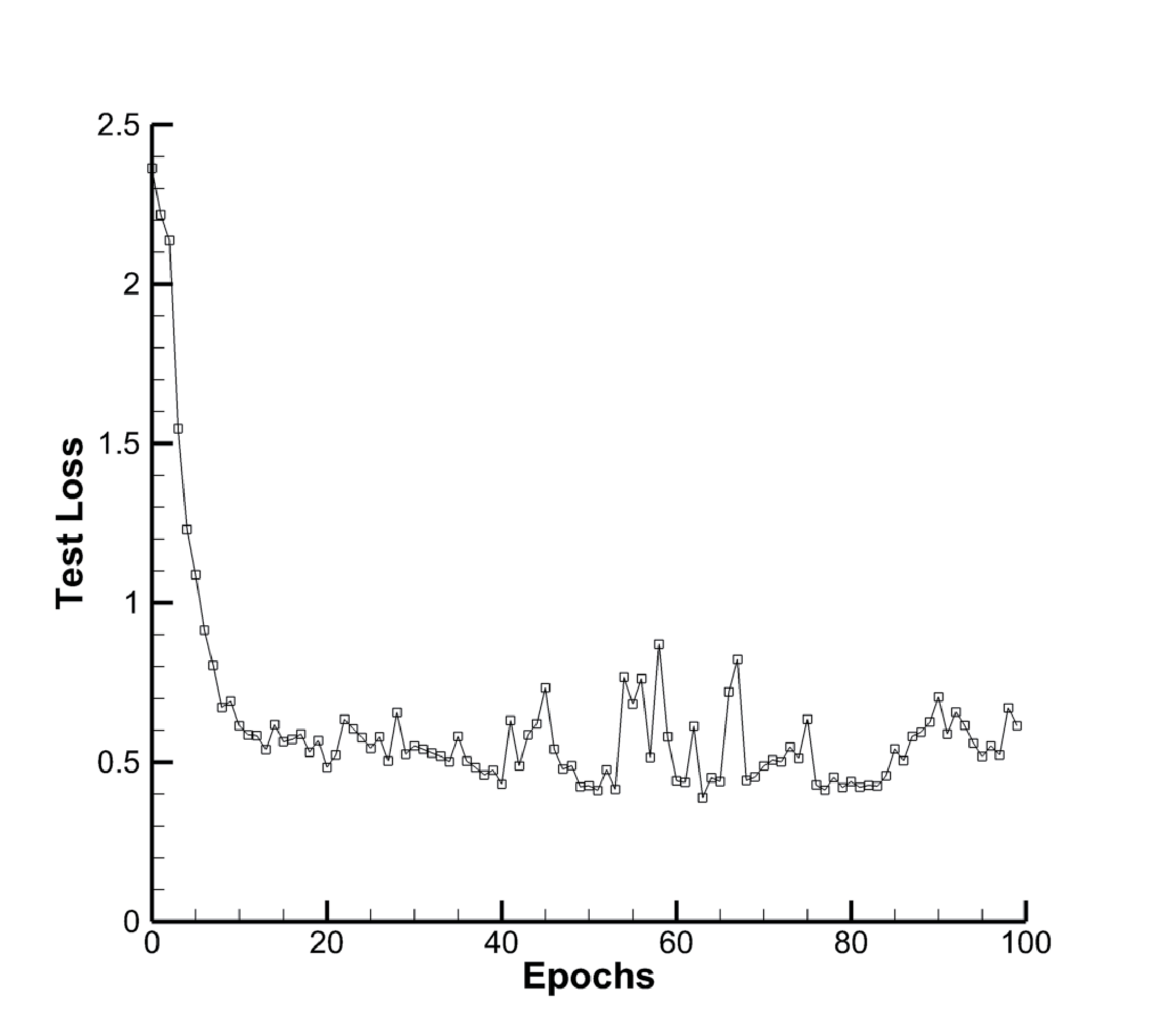}
    \caption{Progression of loss function values (for a language model with parameter count P=60M and training dataset size N=16).}
    \label{fig:result_1}
\end{figure}

\begin{figure}[htbp]
    \centering
    \includegraphics[width=0.9\textwidth]{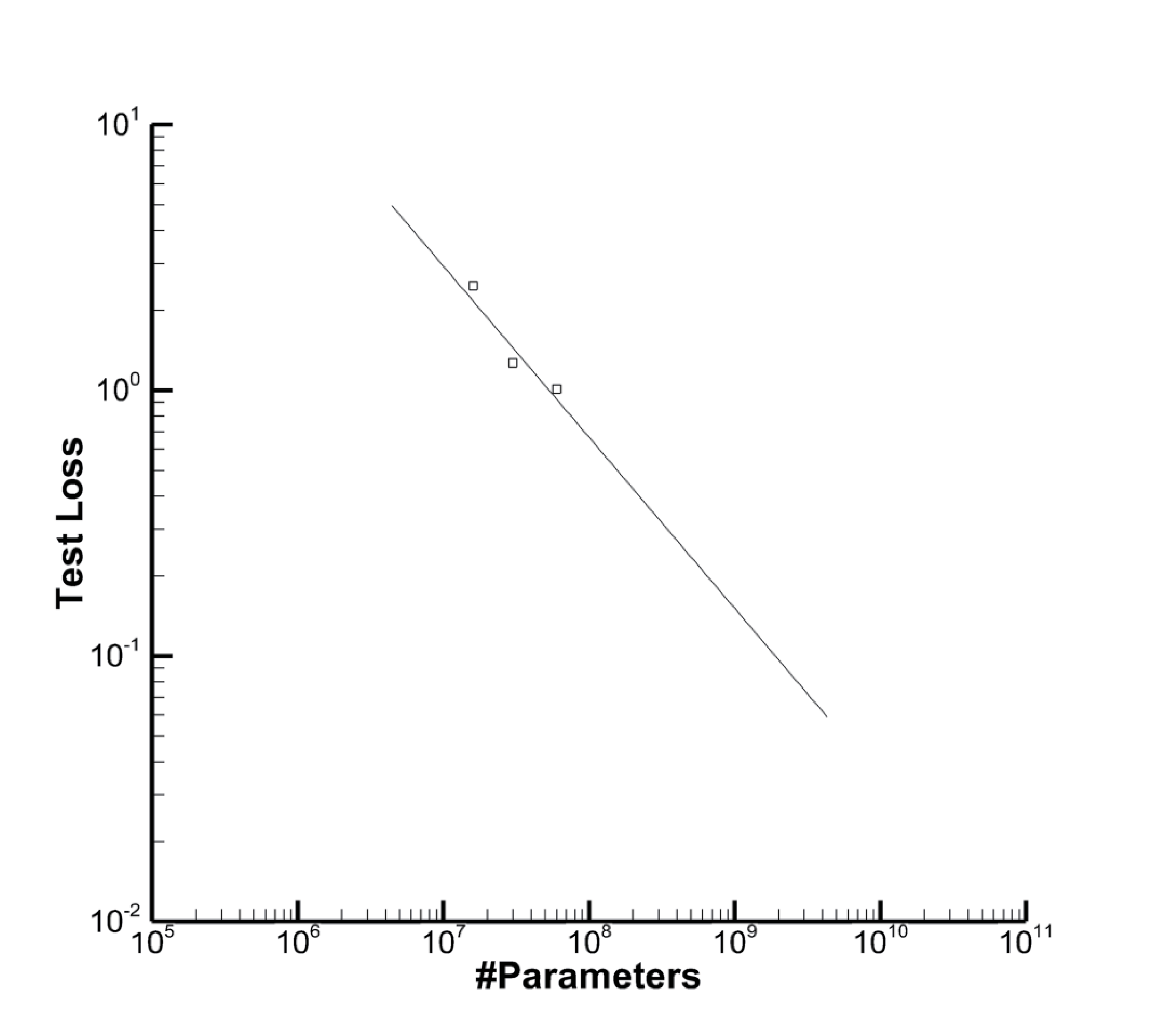}
    \caption{Variation in loss function values (local minima) with the number of parameters (for a training dataset size N=1, after 100 epochs, on a log-log graph).}
    \label{fig:result_2}
\end{figure}

\begin{figure}[htbp]
    \centering
    \includegraphics[width=0.9\textwidth]{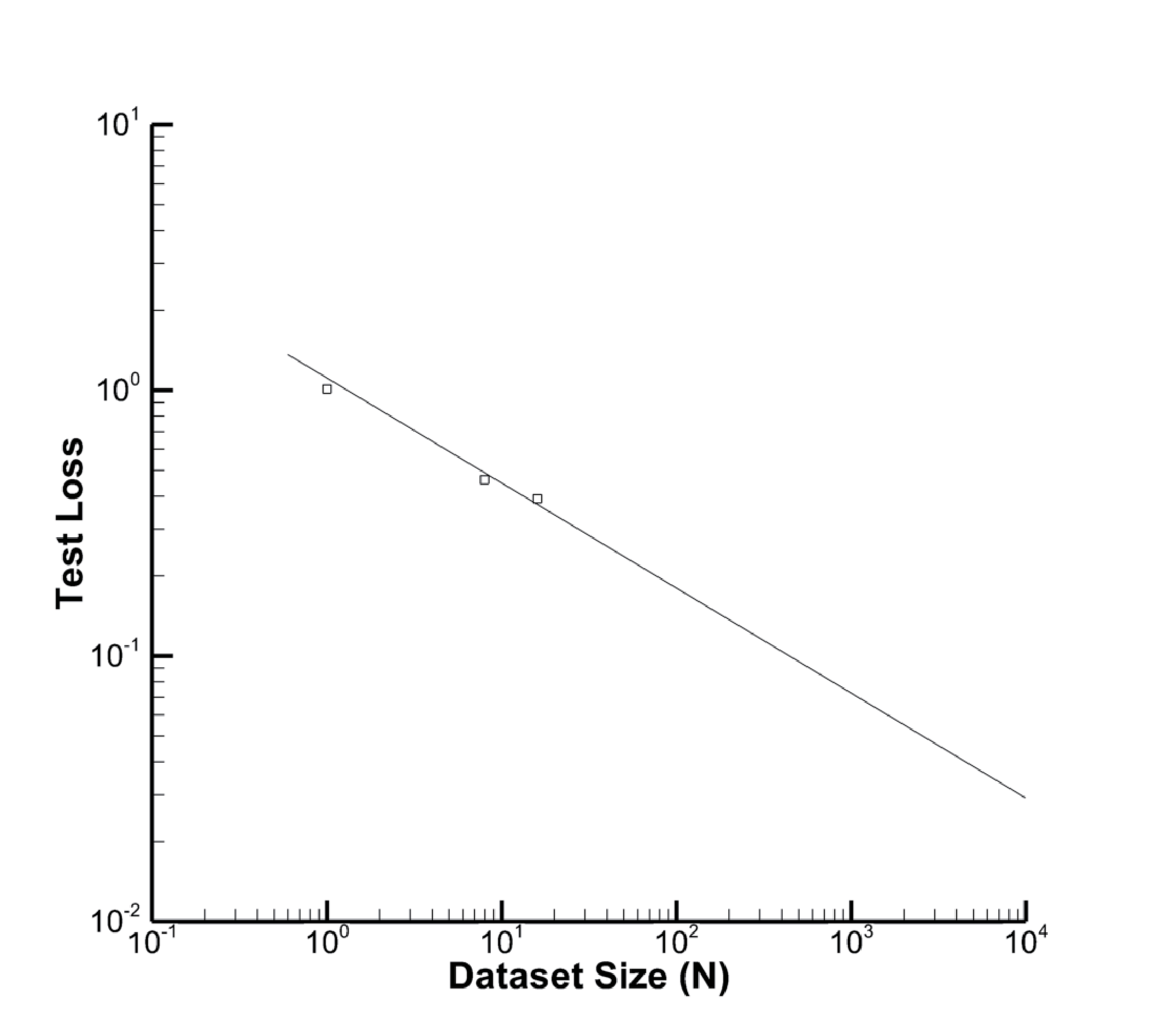}
    \caption{Variation in loss function values (local minima) with the size of the training dataset (for a language model with parameter count P=60M, after 100 epochs, on a log-log graph).}
    \label{fig:result_3}
\end{figure}

Figure \ref{fig:result_1} shows the loss function values (Test Loss) for the test dataset as a function of epoch count, in the case of a language model with parameter count P=60M and training dataset size N=16. Additionally, Figures \ref{fig:result_2} and \ref{fig:result_3} present the loss function values for the test dataset, organized by the two hyperparameters (namely, the number of parameters of the language model and the size of the training dataset).

\section{Discussion}
\subsection{Interpretation of results}
I discuss the results from Figure \ref{fig:result_1}.
Despite changing the substitution rule randomly for each data (text), the Test Loss (representing the cross-entropy error of dictionary reconstruction) is decreasing.
This implies that T5 is capable of learning meta-rules that do not depend on random character substitutions, namely, a general method for deciphering character substitution ciphers. In other words, although the focus is on characters rather than words, T5 demonstrates the ability to acquire some level of equivariance.

From Figures \ref{fig:result_2} and \ref{fig:result_3}, a power law can be discerned.
Readers may recall the power law in previous research \cite{kaplan2020scaling}, where Test Loss due to cross-entropy error was systematically summarized.
Although the scale proposed in this study is also based on cross-entropy error, it differs distinctly as it outputs the results of dictionary estimation.
In other words, while previous research adopted cross-entropy error in the sense of Figure \ref{fig:relationships}b, this study adopts a newly defined cross-entropy error in the sense of Figure \ref{fig:relationships}c.

Regarding Figure \ref{fig:result_2}, although it is an extreme extrapolation and requires further verification, when the Test Loss becomes sufficiently small (here, less than 0.1), the model's number of parameters is estimated to be over 2 billion.
Similarly, the results of Figure \ref{fig:result_3}, also an extreme extrapolation requiring further verification, suggest that for a model with a sufficient number of parameters and after sufficient computation time for training, about N=300, i.e., more than $6.3\times10^8$ tokens of training text, is required for the Test Loss to be sufficiently small.

These results are estimates of the conditions necessary to acquire character-level equivariance and are distinctly different from those needed to acquire word-level equivariance. It is likely that conducting similar numerical experiments at the word level would yield a similar but different power law.

\subsection{Limitations of this study and future directions}
Word substitution ciphers can be interpreted as ciphers with a broader structure than character substitution ciphers. While it is straightforward to extend the model for deciphering character substitution ciphers to word substitution ciphers, there are challenges in its training. Specifically, while the main representations in character substitution ciphers are the 26 letters of the alphabet, the variables in word substitution ciphers correspond to the number of words in the vocabulary, often exceeding 10,000. Therefore, learning their rules is likely to require relatively large computational resources and a large amount of training text.

In this study, after proposing a general theory, I verified through numerical experiments whether T5 could acquire equivariance at the character level. However, it is necessary to test equivariance at the word level using large-scale language models that are publicly available and on par with ChatGPT.

Beyond the scope of this paper, discussing the degree of freedom of the decoded text, that is, the laxity of equivariance in understanding relationships, may contribute to the invention of language models that surpass human or average societal language processing capabilities.

\section{Conclusion}
In this study, I proposed a new hypothesis that hallucinations generated by a language model will disappear if the model sufficiently acquires relationships in real society (A). Next, I explained that the ability to understand text independent of token ID representations can be acquired by the model possessing equivariance (B). Additionally, I proposed a scale to quantify the intensity of hallucinations caused by misunderstandings of relationships, based on cross-entropy error (C). Furthermore, I presented a method for verifying this scale and analyzed it using a toy model (D). Specifically, I measured the deciphering ability of a character substitution cipher by T5, and confirmed that as learning progresses, the value of the loss function incorporating the proposed scale decreases due to the acquisition of equivariance (Figure \ref{fig:result_1}).
Moreover, by varying hyperparameters and conducting numerous numerical experiments, I obtained a power law that is similar to prior research \cite{kaplan2020scaling}, but the derived principles and the results obtained should clearly differ. The extrapolation of the obtained power law revealed that to sufficiently acquire character-level equivariance, a minimum of 2 billion parameters, or texts containing more than $6.3\times10^8$ tokens, are required for training, under the discussion of fixing other hyperparameters.

By scaling up the character-level numerical experiments conducted in this study to the word-level numerical experiments handled by state-of-the-art practical large-scale language models (i.e., learning the rules of deciphering word substitution ciphers and quantifying them with the scale), it is conceivable that we can uniformly discuss the extent to which various types of large-scale language models possess the ability to acquire equivariance that leads to the suppression of hallucinations generated by language models. Furthermore, by extrapolating and utilizing the potentially newly obtained power laws, it can be considered that I will contribute to the discovery and establishment of super large-scale language models freed from hallucinations.

\bibliographystyle{elsarticle-num} 
\bibliography{cas-refs}

\begin{thebibliography}{1}
\expandafter\ifx\csname url\endcsname\relax
  \def\url#1{\texttt{#1}}\fi
\expandafter\ifx\csname urlprefix\endcsname\relax\def\urlprefix{URL }\fi
\expandafter\ifx\csname href\endcsname\relax
  \def\href#1#2{#2} \def\path#1{#1}\fi

\bibitem{devlin2018bert}
J.~Devlin, M.-W. Chang, K.~Lee, K.~Toutanova, Bert: Pre-training of deep bidirectional transformers for language understanding, arXiv preprint arXiv:1810.04805 (2018).

\bibitem{karras2021alias}
T.~Karras, M.~Aittala, S.~Laine, E.~H{\"a}rk{\"o}nen, J.~Hellsten, J.~Lehtinen, T.~Aila, Alias-free generative adversarial networks, Advances in Neural Information Processing Systems 34 (2021) 852--863.

\bibitem{raffel2020exploring}
C.~Raffel, N.~Shazeer, A.~Roberts, K.~Lee, S.~Narang, M.~Matena, Y.~Zhou, W.~Li, P.~J. Liu, Exploring the limits of transfer learning with a unified text-to-text transformer, The Journal of Machine Learning Research 21~(1) (2020) 5485--5551.

\bibitem{tay2021scale}
Y.~Tay, M.~Dehghani, J.~Rao, W.~Fedus, S.~Abnar, H.~W. Chung, S.~Narang, D.~Yogatama, A.~Vaswani, D.~Metzler, Scale efficiently: Insights from pre-training and fine-tuning transformers, arXiv preprint arXiv:2109.10686 (2021).

\bibitem{aldarrab2021can}
N.~Aldarrab, J.~May, Can sequence-to-sequence models crack substitution ciphers?, in: Proceedings of the 59th Annual Meeting of the Association for Computational Linguistics and the 11th International Joint Conference on Natural Language Processing (Volume 1: Long Papers), 2021, pp. 7226--7235.

\bibitem{kambhatla2023decipherment}
N.~Kambhatla, L.~Born, A.~Sarkar, Decipherment as regression: Solving historical substitution ciphers by learning symbol recurrence relations, in: Findings of the Association for Computational Linguistics: EACL 2023, 2023, pp. 2091--2107.

\bibitem{vaswani2017attention}
A.~Vaswani, N.~Shazeer, N.~Parmar, J.~Uszkoreit, L.~Jones, A.~N. Gomez, {\L}.~Kaiser, I.~Polosukhin, Attention is all you need, Advances in neural information processing systems 30 (2017).

\bibitem{kaplan2020scaling}
J.~Kaplan, S.~McCandlish, T.~Henighan, T.~B. Brown, B.~Chess, R.~Child, S.~Gray, A.~Radford, J.~Wu, D.~Amodei, Scaling laws for neural language models, arXiv preprint arXiv:2001.08361 (2020).

\end{thebibliography}
\end{document}